\documentclass[]{bytedance_seed}

\usepackage[toc,page,header]{appendix}
\usepackage{amsfonts}

\usepackage{minitoc}
\usepackage{wrapfig}

\usepackage{graphicx}
\usepackage{capt-of} % 或 \usepackage{caption}

\title{Heptapod: Language Modeling on Visual Signals}

\author[1,*]{Yongxin Zhu}
\author[1]{Jiawei Chen}
\author[1]{Yuanzhe Chen}
\author[1]{Zhuo Chen}
\author[1]{Dongya Jia}
\author[1]{\\Jian Cong}
\author[1]{Xiaobin Zhuang}
\author[1]{Yuping Wang}
\author[1]{Yuxuan Wang}

\affiliation[1]{ByteDance Seed}

\contribution[*]{Work done during an internship at ByteDance Seed}
\abstract{
We introduce Heptapod\footnotemark, an image autoregressive model that adheres to the foundational principles of language modeling. Heptapod employs \textbf{causal attention}, \textbf{eliminates reliance on CFG}, and \textbf{eschews the trend of semantic tokenizers}. Our key innovation is \textit{next 2D distribution prediction}: a causal Transformer with reconstruction-focused visual tokenizer, learns to predict the distribution over the entire 2D spatial grid of images at each timestep. This learning objective unifies the sequential modeling of autoregressive framework with the holistic self-supervised learning of masked autoencoding, enabling the model to capture comprehensive image semantics via generative training. On the ImageNet generation benchmark, Heptapod achieves an FID of $2.70$, significantly outperforming previous causal autoregressive approaches. We hope our work inspires a principled rethinking of language modeling on visual signals and beyond.
}

\date{\today}
\correspondence{Yongxin Zhu at \email{yongxin.zhu@bytedance.com}, Zhuo Chen at \email{zhuo.chen1@bytedance.com}}

\begin{document}
\maketitle

\footnotetext{Heptapod refers to the alien species in the film \textit{Arrival} (2016). Their circular logograms encode complete messages holistically and reflect a non-linear perception of time. This instantaneous, atemporal writing system parallels the core idea of our framework.}

%不需要目录就注释掉 注意目录不要和第一页放在一块 要有\newpage
%\newpage
%\tableofcontents
%\newpage

\section{Introduction}

\begin{figure}[t]
  \centering
  \includegraphics[width=1.0\linewidth]{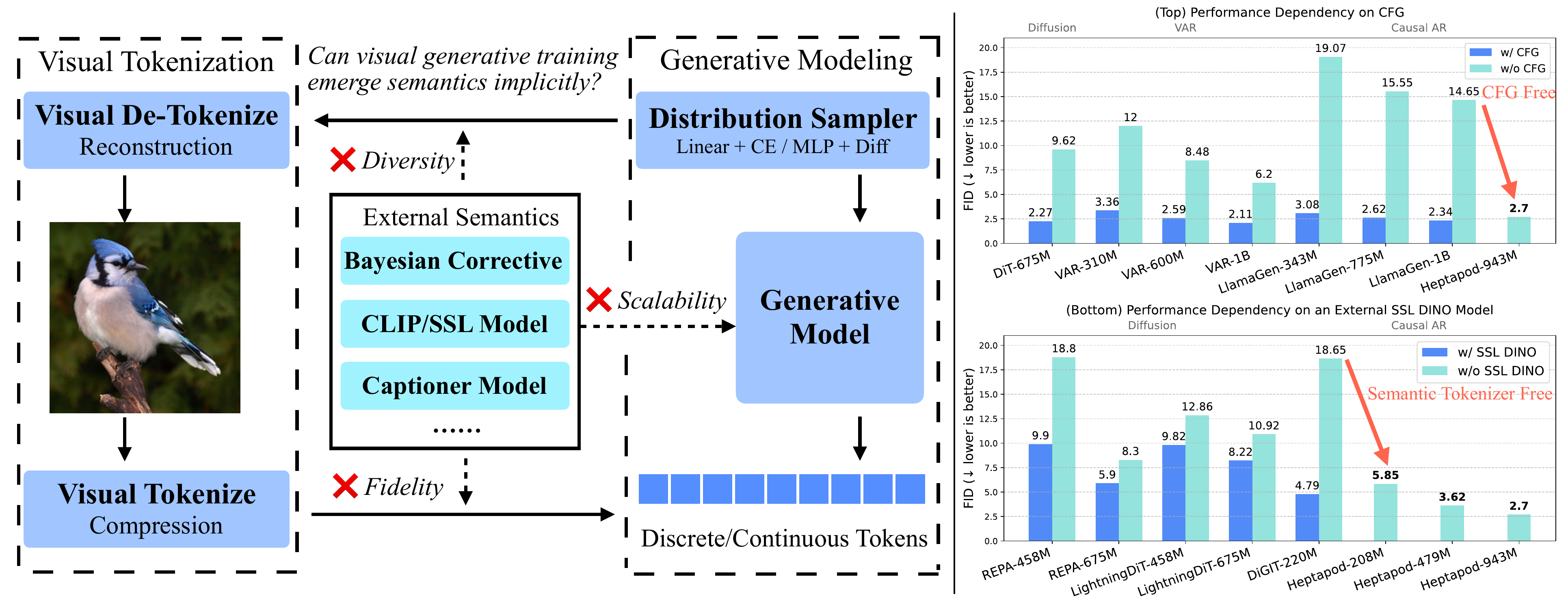}
  \caption{(\textbf{Left}) A typical visual latent generative framework that incorporates external semantics. (\textbf{Top Right}) Leading autoregressive models exhibit steep performance drops when CFG is disabled (VAR results from \citet{chen2025toward}). (\textbf{Bottom Right}) Generation quality degrades when the external SSL model DINO is removed (CFG is disabled).}
  \label{fig:intro1}
\end{figure}

The emergence of Large Language Models (LLMs) \cite{radford2018improving, radford2019language, brown2020language} has precipitated a paradigm shift in artificial intelligence. Their success is widely attributed to a simple yet powerful recipe: a scalable Transformer with causal attention \cite{vaswani2017attention}, an efficient BPE tokenizer for character-level compression \cite{sennrich2016neural}, and a straightforward self-supervised objective of next-token prediction. This formula has catalyzed a surge of work that seeks to transplant these principles to other modalities \cite{pmlr-v119-chen20s,sun2024autoregressive,NEURIPS2024_9a24e284,NEURIPS2024_325ce329,lakhotia-etal-2021-generative,nguyen-etal-2025-spirit}. As illustrated in Fig.~\ref{fig:intro1} (Left), a typical visual generation framework mirrors this structure: a tokenizer compresses high-dimensional pixels into a latent space, and a generative model learns the distribution over the resulting representations. However, directly transferring the language modeling paradigm from one-dimensional text to the two-dimensional visual domain has proven challenging, which has prompted many approaches to incorporate what we term external semantics—information or guidance not learned from the next-token prediction objective—to bridge the performance gap.

A primary manifestation of this dependency is the heavy reliance on Classifier-Free Guidance (CFG) \cite{ho2021classifierfree}, an inference-time technique that refines the generated distribution by bayesian correction. While effective, reliance on this external corrective mechanism not only obscures intrinsic limitations of the model's ability in learning from visual signals, but also leads to intensity oversaturation and reduced sample diversity. As shown in Fig.~\ref{fig:intro1} (Top Right), the performance of leading visual generative models \cite{Peebles_2023_ICCV,NEURIPS2024_9a24e284,sun2024autoregressive} degrades substantially when CFG is disabled, highlighting a dependence on this external crutch. This pattern suggests that the self-sufficient language modeling paradigm has not yet cleanly translated to visual signals.

Another popular strategy embeds external semantics directly into the tokenizer, treating it as the linchpin for success \cite{yu2024language,NEURIPS2024_e91bf7df,Han_2025_CVPR}. Inspired by semantic tokenizers in speech \cite{lakhotia-etal-2021-generative,lee-etal-2022-textless,NEURIPS2023_c859b99b,nguyen-etal-2025-spirit,nguyen-etal-2023-generative}, these approaches learn visual vocabularies by distilling from pretrained self-supervised learning (SSL) models \cite{NEURIPS2024_325ce329,Qu_2025_CVPR,li2025imagefolder,ma2025unitok,geng2025xomni}. As demonstrated in Fig.~\ref{fig:intro1} (Bottom Right), incorporating an external SSL model such as DINO \cite{oquab2024dinov} can markedly improve image generation. However, we contend that this departs from the principles that underpin LLM success. The BPE tokenizer is semantically agnostic and its sole purpose is faithful data compression. The semantic relationships are not engineered into the tokenizer but rather emerge within the Transformer under the next-token prediction objective. By analogy, constructing a semantic tokenizer for text from pretrained embeddings such as GloVe \cite{pennington-etal-2014-glove} or BERT \cite{devlin-etal-2019-bert} would run counter to the very principles that made LLMs successful. Moreover, in audio, semantic tokenizers are known to sacrifice fidelity (e.g., acoustic detail) \cite{kreuk-etal-2022-textless,zhang2024speechtokenizer}. This suggests a dilemma of ``impossibility triangle'' between reconstruction, generation, and semantic/SSL representation \cite{ramanujan2024worse,Yao_2025_CVPR,yu2024language}, where optimizing one objective can compromise the others. This introspection leads to the central question of our work:

\textit{Can we devise a visual generative learning paradigm that returns to first principles, where the tokenizer is dedicated solely to faithful reconstruction, and complex semantics emerge implicitly within the Transformer through the next-token prediction objective?}

\begin{figure}[t]
  \centering
  \begin{minipage}[c]{0.57\textwidth}
    \centering
    \includegraphics[width=\linewidth]{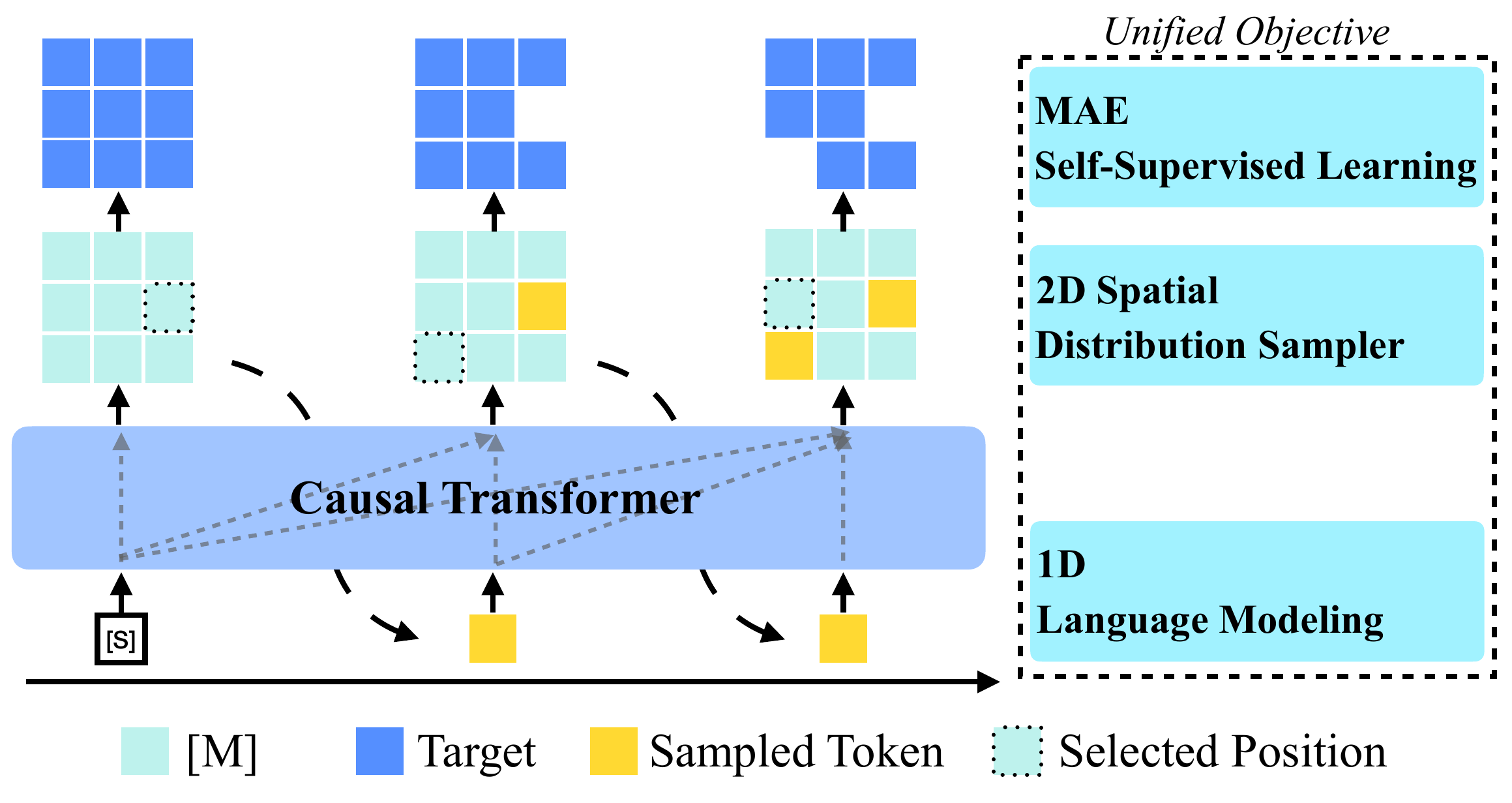}
  \end{minipage}\hfill
  \begin{minipage}[c]{0.42\textwidth}
    \captionof{figure}{Illustration of Heptapod's next 2D distribution prediction framework. The model operates on a sequence of visual tokens from a simple reconstruction-focused tokenizer. The Transformer autoregressively predicts the distributions over remaining positions in the 2D grid in parallel for every input tokens. The loss is then computed across all these future positions, treating the prefix as the visible context (like MAE's unmasked patches) and the remaining grid as targets. This forces the model to develop a holistic representation, bridging the gap between 1D language modeling and 2D spatial understanding.}
    \label{fig:intro2}
    \small
  \end{minipage}
\end{figure}

Addressing this question requires confronting a core ambiguity: the notion of ``next token'' is ill-defined in two-dimensional space. Unlike text, which has a natural 1D temporal order, images lack an intrinsic sequence. Any patch in a 2D grid could be considered ``next'' by spatial proximity or semantic relatedness (e.g., two eyes in a portrait). To resolve this, we introduce \textbf{Heptapod}, a framework that generalizes next-token prediction from a 1D sequence to a holistic 2D distribution prediction. As illustrated in Fig.~\ref{fig:intro2}, our framework consists of a standard causal Transformer that ingests visual tokens produced by a reconstruction-focused tokenizer (such as VQ-VAE \cite{NIPS2017_7a98af17} or VAE \cite{Kingma2013AutoEncodingVB}). Unlike traditional autoregressive models that predict the token at a \textit{single designated} next position, our model is trained to predict, in parallel, the distribution of the token at \textit{every} subsequent spatial position in the image. This design enables the model to capture complex spatial dependencies and holistic image semantics, rather than relying on a hand-crafted order. In this framework, discrete token prediction is handled via a linear classification head with cross-entropy loss, whereas continuous token prediction employs a diffusion-style head \cite{NEURIPS2024_66e22646} with MSE loss, both viewed as instances of modeling distributions over latent visual representations.

By training a Transformer with 1D causal attention to predict a full 2D distribution, we compel the model to develop a holistic understanding of the image. To accurately predict the distribution at any position, the Transformer must encode a compact, predictive representation of the image's structure and semantics. Furthermore, from a self-supervised learning perspective, our objective unifies autoregressive modeling with Masked Autoencoding (MAE) \cite{He_2022_CVPR}: the causal prefix serves as the unmasked context, while predicting the entire 2D grid in parallel is analogous to reconstructing masked patches. In summary, our contributions are:
\begin{itemize}
\item We challenge the prevailing trend of incorporating external semantics into visual generative models, advocating for a paradigm that decouples reconstruction (tokenizer) from semantic learning (Transformer).
\item We introduce next 2D distribution prediction, a novel objective that generalizes autoregressive modeling to non-sequential data by reformulating next-token prediction as a holistic spatial task.
\item We provide a unified perspective that integrates the core principle of MAE-style SSL into the next-token prediction paradigm in a causally coherent manner.
\end{itemize}

\section{Related Work}

\subsection{Visual Tokenization}

Visual tokenization for language modeling generally follows two distinct philosophies. The first is grounded in the principle of reconstruction, aiming to produce a compressed yet faithful representation of an image. This line was pioneered by models such as VQ-VAE \cite{NIPS2017_7a98af17} and VAE \cite{Kingma2013AutoEncodingVB}, with subsequent advances improving perceptual quality and rate-distortion trade-offs. For example, VQGAN \cite{Esser_2021_CVPR} introduced perceptual and adversarial losses to specifically enhance reconstruction fidelity. More recent research on VQ-VAE has focused on addressing codebook collapse to further improve reconstruction quality, gradually bridging the gap to continuous VAE \cite{zhu2024addressing,mentzer2024finite,yu2022vectorquantized,yu2024language,Zhu2024ScalingTC}. Despite these methods yield powerful tokenizers for image compression and reconstruction, a persistent challenge remains: autoregressive models trained on these tokenizers often struggle to learn visual signals, resulting in generative quality heavily dependent on CFG \cite{ho2021classifierfree}.

These limitations motivated a second philosophy, inspired by the semantic tokenizer in speech generation \cite{lakhotia-etal-2021-generative}. In this vein, recent approaches \cite{NEURIPS2024_325ce329,Yao_2025_CVPR,leng2025repa,li2025imagefolder,Qu_2025_CVPR,ma2025unitok} distill prior knowledge from pre-trained SSL models such as DINO \cite{Caron_2021_ICCV,oquab2024dinov} or CLIP \cite{radford2021learning,Zhai_2023_ICCV}, either to construct a ``semantic'' vocabulary or to inject semantic information directly into the generative model. However, this strategy is inherently constrained by the capabilities of  external SSL models and often suffers from information loss that impairs reconstruction quality \cite{kreuk-etal-2022-textless}. Moreover, joint optimization of reconstruction and semantic objectives often reveals a fundamental tension between these competing goals \cite{Yao_2025_CVPR}. 
In contrast, we sidestep this dilemma. We adopt a straightforward reconstruction-based tokenizer but fundamentally alter the learning objective. By introducing a next-2D distribution prediction objective, we enable the Transformer to learn visual semantics directly, relieving the tokenizer of this burden and preserving its focus on faithful compression and reconstruction.

\subsection{Image Language Modeling}

The application of autoregressive language modeling to the visual domain began with pioneering works such as iGPT \cite{pmlr-v119-chen20s}, which operated directly on pixels but did not achieve the generative prowess observed in text-based language models. Subsequent work \cite{Esser_2021_CVPR,yu2022vectorquantized,sun2024autoregressive,yu2024randomized} combined VQGAN tokenizers with Transformers, establishing a dominant paradigm for visual autoregressive generation. However, when trained with a standard next token prediction objective, these models typically require CFG \cite{ho2021classifierfree} to produce competitive samples, indicating that a direct transfer of language modeling methodology from text to visual signals does not fully address the challenge of modeling complex visual distributions.
In parallel, alternative approaches have diverged from the classic autoregressive framework. For instance, MAR \cite{NEURIPS2024_66e22646} proposes a generalized autoregressive framework but abandons the causal attention mechanism. Similarly, VAR \cite{NEURIPS2024_9a24e284} performs coarse-to-fine autoregression but employs non-causal attention within each scale. Both methods depart from the causal structure characteristic of LLMs, which we seek to preserve and extend.
Our work charts a distinct course. We retain the causal attention mechanism and utilize a simple reconstruction-based tokenizer. By reformulating next token prediction as holistic next 2D distribution prediction, we force the causal Transformer to develop a global understanding of the visual space. This approach bridges the gap between local sequential processing and holistic spatial generation, adhering to the first principles that underpin successful language modeling.

\section{On the Nature of Visual Tokens and Their Predictive Modeling}
\label{sec:learn_objective}

Adapting the language modeling paradigm to the visual domain necessitates a foundational choice regarding the nature of a visual ``token'' and the objective function used for its prediction. Unlike text, where discrete tokens are naturally suited to sequential modeling, visual signals can be represented either as discrete latent codes from VQ-VAE or as continuous latent vectors from VAE. Historically, the choice of visual tokenizer was driven by the superior reconstruction fidelity of VAE. However, recent work \cite{weber2024maskbit} has demonstrated that VQ-VAE models with sufficiently large codebooks can match or even surpass VAE, thereby narrowing the distinction between these two tokenization strategies.

In this work, we step back from this ongoing debate and argue that from the perspective of language modeling, the distinction between discrete and continuous tokens is mathematically immaterial. The core task of language modeling is to model the distribution of the next token $z_t$, conditioned on the context encoded in the Transformer's hidden state $h_{t-1}$. The choice of token simply dictates the methodology used to parameterize and optimize this distribution, but the objective of accurate conditional modeling remains the same.
For discrete tokens, the distribution is a categorical one over a finite vocabulary, which can be directly parameterized with a softmax layer and optimized via cross-entropy loss. Minimizing this loss is equivalent to directly maximizing the log-likelihood of the observed token sequence, providing a straightforward and powerful learning signal.
The modeling of continuous tokens is more complex as their distributions typically defy representation by a simple parametric form. To address this, diffusion is adopted as it can model complex non-parametric distributions \cite{NEURIPS2024_66e22646}. The diffusion objective is typically an MSE loss on the noise or the original token, which has been shown equivalent to maximizing a lower bound on the log-likelihood of the data \cite{kingma2021variational, ho2020denoising}.

Although the probabilistic objectives align, their distinct parameterizations have significant implications for training dynamics, which our experiments corroborate in Sec. \ref{sec:tokenizer}. The cross-entropy loss provides a direct and sharply defined gradient signal such that the model receives unambiguous feedback distinguishing the ground truth token from all other distractors. In contrast, the implicit learning signal provided by the diffusion loss is averaged over a continuum of noise levels. While effective for learning smooth densities, it dilutes the gradient signal and potentially slows the convergence.

\section{Next 2D Distribution Prediction}

In the above discussion, we established that whether discrete or continuous visual tokens, the learning objective can be unified as modeling the token distribution. However, the results presented in Fig. \ref{fig:intro1} show that a naive next-token prediction objective applied to visual tokens yields unsatisfactory performance. This raises a fundamental question: for a two-dimensional image, what does ``next token'' actually mean? In this section, we analyze the root of this problem and introduce our solution next 2D distribution prediction.

\begin{figure}[t]
  \centering
  \includegraphics[width=1.0\linewidth]{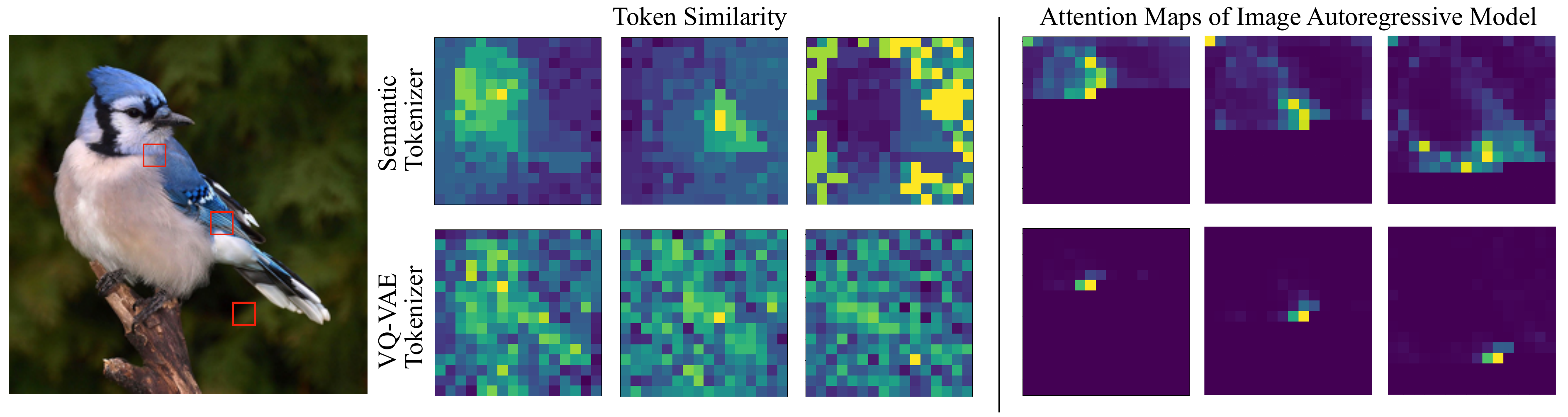}
  \caption{\textbf{(Left)} Spatial correlations in VQ-VAE vs.\ semantic tokenizers. For three reference tokens (87th, 138th and 203rd, bounded by red lines), we compute cosine similarity to all other tokens in the grid. \textbf{(Right)} Attention maps of the final layer in autoregressive Transformer trained with each tokenizer. Under VQ-VAE, attention concentrates on spatial neighbors (local interpolation), while semantic tokens yield attention on spatially distant yet semantically related regions (long-range dependencies). Following DiGIT \cite{NEURIPS2024_325ce329}, semantic tokens are obtained by K-Means on DINO hidden states. Additional examples are provided in Appendix~\ref{appendix:similarity_grid}.}
  \label{fig:token_visualization}
\end{figure}

\subsection{The Curse of Locality in Visual Autoregressive Models}

The struggles of language modeling on visual signals stem from a fundamental difference between text and images in information density and spatial correlation. In the textual domain, language is highly abstract and compressed. Predicting the next token requires understanding long-range grammatical structures and semantic dependencies beyond local patterns. Consequently, high-level semantics emerge naturally as a byproduct of optimizing next-token prediction. Images, by contrast, are highly redundant \cite{He_2022_CVPR}. As illustrated in Fig.~\ref{fig:token_visualization}, VQ-VAE tokens exhibit strong local correlations that neighboring tokens are overwhelmingly similar. Under teacher forcing with a fixed scan order, a visual autoregressive model quickly discovers a shortcut to excel at local interpolation. It can substantially reduce loss by perfectly predicting adjacent, highly correlated tokens, with little incentive to capture the long-range dependencies that are crucial for global structure but provide only marginal additional loss reduction. Optimization thus gravitates to a local minimum that favors textures and low-level detail over holistic semantics.

This precisely explains why semantic tokenizers often boost performance. SSL models are typically forced to learn long-range denpendencies via contrastive learning \cite{Caron_2021_ICCV} or high-ratio masked prediction \cite{He_2022_CVPR}. By distilling knowledge from SSL models, semantic tokens, as shown in Fig. \ref{fig:token_visualization}, pre-package long-range semantic relationships into the tokens themselves. The autoregressive model is then forced to learn these pre-computed global dependencies. The attention maps show that the autoregressive model attents to the spatially distant yet semantically related regions, revealing the long-range dependencies. However, this workaround departs from the first principles of language modeling, where semantics should emerge inside the Transformer from the learning objective rather than be engineered into the tokenizer. This compromise not only hurts reconstruction fidelity \cite{kreuk-etal-2022-textless,Yao_2025_CVPR} but also limits the model's ability to learn knowledge beyond that of the pretrained SSL model. Our work aims to break this impasse to design a learning paradigm that retains the fidelity of a reconstruction-based tokenizer while intrinsically compelling the Transformer to learn global semantics.

Motivated by the semantic tokenizer, we redefine ``next token'' for visual data to explicitly target long-range dependencies. At each timestep, rather than predicting a single next token at a specified location, the model predicts the distribution over the tokens at all positions in the 2D spatial grid given the current causal prefix, eliminating the local interpolation shortcut. To accurately predict spatially nonadjacent patches, the model must infer the global structure and semantics from the visible context. In this new paradigm, understanding global semantics is no longer an option, but a necessity for optimizing the objective.

\subsection{The 2D Prediction Objective}

\begin{figure}[t]
  \centering
  \includegraphics[width=1.0\linewidth]{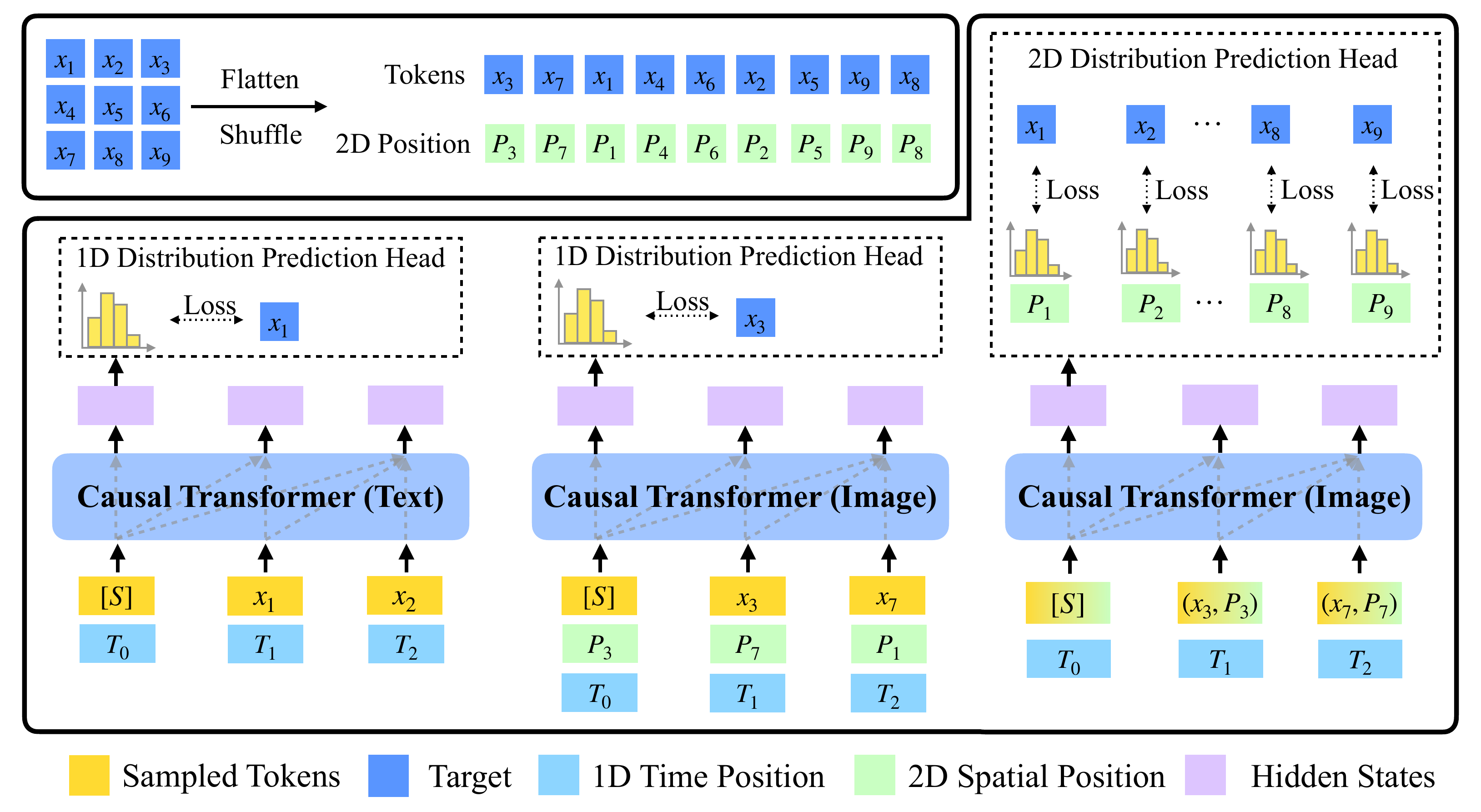}
  \caption{\textbf{(Left)} The text language modeling with next 1D distribution prediction. \textbf{(Middle)} Vanilla image language modeling with next 1D distribution prediction. The 2D sptial postion is left-shifted in the input to specify the next target. \textbf{(Right)} Our next 2D distribution prediction. The 2D spatial positions are not shifted. The model must be prepared to predict any future position.}
  \label{fig:model_arch}
\end{figure}

The core mechenism of next 2D distribution prediction is to elevate the autoregressive target from a 1D vocabulary to a 2D vocabulary (the distribution over the entire image grid), while fully preserving the Transformer's 1D causal attention mechanism. To understand this, we compare this paradigm shift in Fig. \ref{fig:model_arch}. In text, the notion of ``next'' is defined by the 1D temporal order. The model receives the token $x_t$ and its 1D time position $T_t$ at the timestep $t$ and its task is to predict the token distribution for the timestep $t+1$. The model does not need to explicitly decide which position to predict, as it is always the subsequent one. For images, the next token could belong to any of the remaining spatial positions. To resolve this, vanilla image autoregressive models must explicitly be told which spatial position to predict next by left-shifting the 2D position sequence. This forces the 1D causal attention to reason on the 2D spatial space explicitly. Instead, our next 2D distribution prediction framework treats the current token and its own 2D spatial position as ``one token'' from a 2D vocabulary. The model is given no information about which spatial position it should predict next so that it must be prepared to predict any of them.

From a language modeling perspective, next 2D distribution prediction expands the vocabulary from a 1D token space to a 2D (position, token) space. yet the underlying Transformer remains a standard 1D causal model. This design offloads the complexity of modeling 2D space from the causal Transformer to a specialized prediction head, encouraging the model to learn global semantics and long-range dependencies without altering its causal nature. During inference, the model predicts a distribution over the 2D grid at each step, then samples a (position, token) pair to serve as the input for the next timestep util the grid is filled.

\subsection{Architectures for 2D Prediction Head}
\label{sec:arch_design}

\begin{figure}[t]
  \centering
  \includegraphics[width=1.0\linewidth]{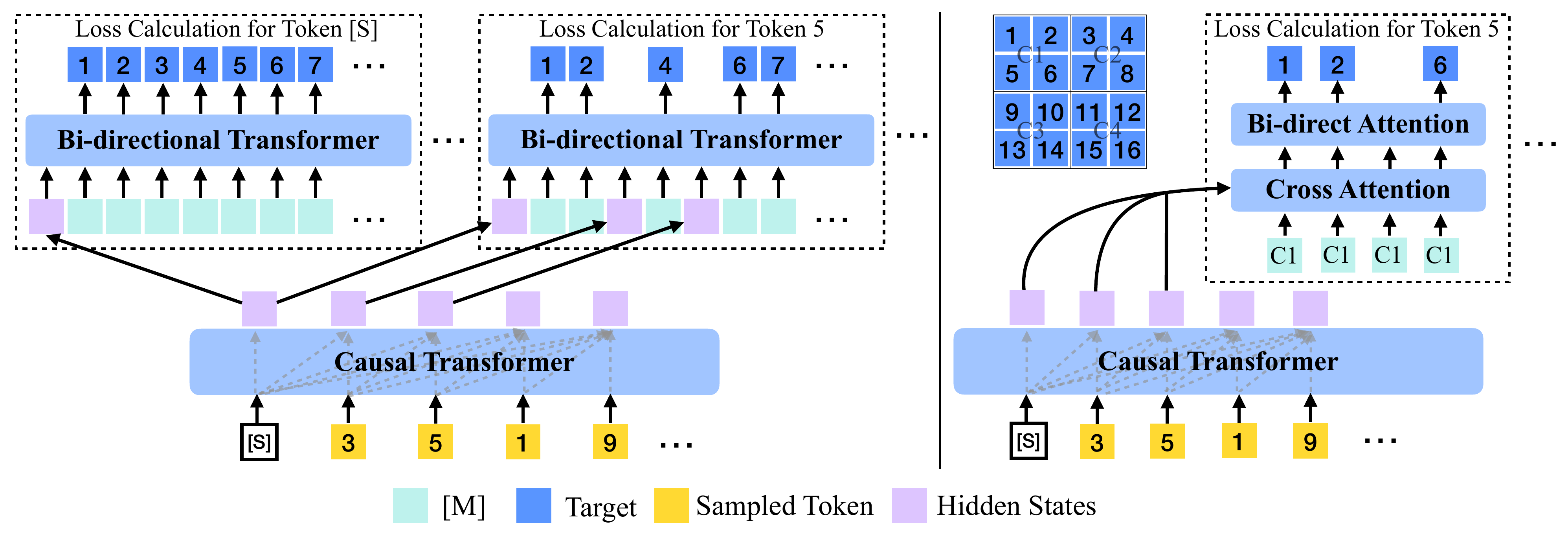}
  \caption{Architectural variants of the 2D prediction head. \textbf{(Left)} The global prediction head employs a bidirectional transformer to model full-image spatial dependencies. \textbf{(Right)} The local prediction head stacks cross attention and bidirectional attention to capture 2D spatial dependencies within a local chunk.}
  \label{fig:lm_head}
\end{figure}

The core of next 2D distribution prediction framework is a prediction head that maps the 1D causal Transformer's output to a full 2D spatial distribution. We explore two architectural variants that trade off modeling scope and computational cost, as illustrated in Fig. \ref{fig:lm_head}.

The first variant, depicted in the left of Fig. \ref{fig:lm_head}, models global spatial dependencies across the entire image grid. At timestep $t$, the hidden states from the causal Transformer $h_{1:t}$ represent the unmasked portion of the image. To predict the complete 2D grid, we combine $h_{1:t}$ with a set of learnable mask tokens, each augmented with a learnable spatial positional embedding corresponding to a distinct masked position ($t+1$ to $N$). The concatenated sequence is then processed by a bi-directional Transformer, allowing every position to attend to every other. The model is enforced to explicitly reason about global spatial relationships and infer the content of the masked regions based on the visible context. The loss is computed only on outputs associated with masked positions, forcing the model to perform a holistic prediction of the unseen future from the causal past. While this global head captures long-range dependencies effectively, its computational cost scales with the grid size.

To improve efficiency, we design a second variant that operates on local regions, as shown in the right of Fig. \ref{fig:lm_head}. We partition the image into non-overlapping chunks. At each timestep, the prediction head focuses only on the chunk containing the next token to be predicted. The head interleaves cross-attention and bidirectional attention layers. Learnable queries, corresponding to positions within the target chunk, first attend to the historical context from the causal Transformer's hidden states via cross-attention. Subsequently, bidirectional self-attention layers operate exclusively within the chunk, modeling local spatial dependencies in that specific region. This design reduces the computational complexity from the scale of the entire image to that of a single chunk.

Both architectures, whether global or local, implement our framework's core principle: a single 1D causal step triggers a comprehensive 2D spatial prediction. This mechanism compels the model to develop a holistic understanding of visual data within a standard autoregressive training process.

\section{Experiments}

\subsection{Experimental Setup}

We conduct autoregressive image generation experiments on ImageNet-1K \cite{imagenet} dataset at a resolution of $256\times256$. We report FID \cite{NIPS2017_8a1d6947} and IS \cite{NIPS2016_8a3363ab} as metrics. To assess intrinsic generative capability and ensure fair comparison, we disable CFG in all experiments.

We evaluate our framework with both discrete and continuous visual tokenizers as discussed in Sec.~\ref{sec:learn_objective}. For discrete tokenization, we use the VQGAN tokenizer from LlamaGen \cite{sun2024autoregressive}, which attains a rFID of $2.19$. For continuous tokenization, we use the VAE tokenizer from MAR \cite{NEURIPS2024_66e22646}, which achieves a rFID of $1.43$. Both tokenizers compress a $256\times256$ image into a $16\times16$ latent grid, yielding a sequence of $256$ tokens. For continuous tokens, we follow MAR \cite{NEURIPS2024_66e22646} and adopt a diffusion-style MLP layer.

Unless otherwise stated, we follow MAR \cite{NEURIPS2024_66e22646} and match the size of the prediction head to the causal Transformer, including hidden width, number of layers, number of attention heads, and the MLP expansion ratio. The local prediction head contains one additional attention block per layer compared to the global variant. We provide the model configurations and ablations on the prediction head in Appendix \ref{appendix:configuration}. All models are trained for $800$ epochs on the ImageNet training split with a batch size of $2048$ and learning rate of $8e-4$, using AdamW with $\beta_1=0.9$ and $\beta_2=0.95$.

\subsection{Convergence Efficiency with Different Tokenizers}
\label{sec:tokenizer}

\begin{figure}[t]
  \centering
  \includegraphics[width=1.0\linewidth]{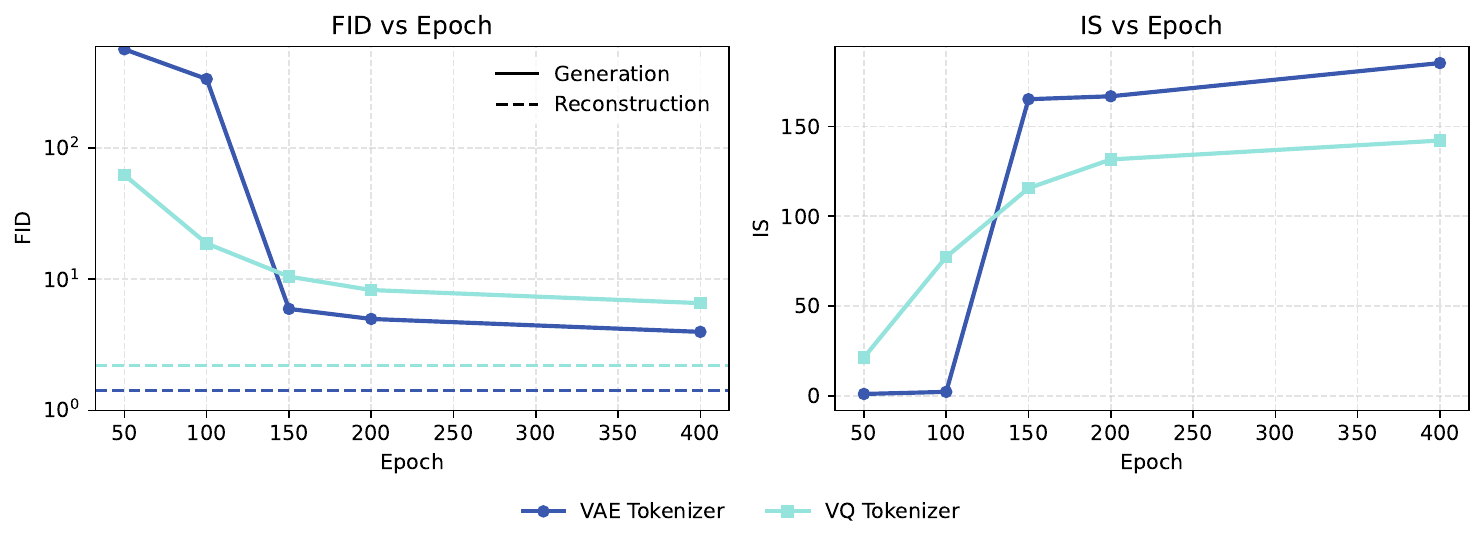}
  \caption{Training convergence for Heptapod-L with a discrete VQ tokenizer vs.\ a continuous VAE tokenizer. The VQ-based model converges faster initially, while the VAE-based model ultimately attains superior generative performance. The dotted line indicates each tokenizer's rFID.}
  \label{fig:fid_is_side}
\end{figure}

As discussed in Sec. \ref{sec:learn_objective}, our framework can accommodate both discrete and continuous visual tokens, as both ultimately aim to maximize the data log-likelihood. To empirically investigate the practical implications of this choice, we train two models of identical scale (Heptapod-L) using the next 2D distribution prediction objective. The first model utilizes a discrete VQ tokenizer with cross-entropy loss to learn the distribution, while the second employs a continuous VAE tokenizer paired with a diffusion MLP head \cite{NEURIPS2024_66e22646}. The results reveal a trade-off between convergence speed and final generative quality.

As shown in Fig. \ref{fig:fid_is_side}, the training dynamics of the two models differ significantly. The model trained with the discrete VQ tokenizer exhibits markedly faster and smoother initial convergence, achieving lower FID and higher IS in early epochs. This supports the hypothesis that cross-entropy provide a more efficient optimization signal. However, the performance of the VAE-based model improves dramatically after $\sim150$ epochs, ultimately surpassing the VQ-based model. This eventual superiority aligns with the underlying reconstruction fidelity of the tokenizer itself. Notably, the final performance gap in generative quality between the two models closely mirrors this gap in their reconstruction capabilities.

The results lead to two main conclusions. First, our experiments validate that reconstruction-focused tokenizers, whether discrete or continuous, provide a viable foundation for language modeling with next 2D distribution prediction. Both approaches enable the model to converge effectively and achieve strong generative results. However, we observe that the reconstruction quality of the tokenizer sets a practical upper bound on the final generative performance of the autoregressive model. Given its superior performance upon full convergence, we adopt the VAE tokenizer as our default choice for subsequent experiments.
Second, the choice of tokenizer and its corresponding loss function has a significant impact on training efficiency, which may become a factor when scaling up the model. This analysis raises an intriguing possibility: a discrete VQ tokenizer, if its reconstruction quality were improved to match or exceed that of the VAE, could potentially offer the best of both worlds, combining the rapid convergence of discrete optimization with generative performance of a continuous VAE.

\subsection{Properties of Next 2D Distribution Prediction}
\label{sec:properties}

To characterize the properties of the next 2D distribution prediction objective, we conduct an ablation on the prediction head design. Table~\ref{tab:order} shows that randomizing the generation order in a standard 1D next-token paradigm (1D-random) improves over a fixed raster scan (1D-raster), but Heptapod (2D-random) delivers substantially larger gains. This suggests that explicitly predicting distributions over the 2D images is more impactful than the 1D next-token prediction on visual signals.

\begin{wraptable}{r}{0.45\linewidth}
  %\vspace{\baselineskip} 
  \centering
  \caption{Impact of distribution sampler (1D vs.\ 2D) and generation order. A large size model is trained for $400$ epochs with CFG disabled. 1D-raster predicts a single next token in a fixed raster scan. 1D-random predicts a single next token at uniformly random spatial positions. 2D-random (Heptapod) predicts distributions over all spatial positions and samples a (position, token) pair each step.}
  \label{tab:order}
  \begin{tabular}{lcr}
      \toprule
      Sampler \& Order & FID$\downarrow$ & IS$\uparrow$ \\
      \midrule
      1D-raster & 19.23 & 62.3 \\
      1D-random & 13.07 & 91.4 \\
      2D-random (Heptapod) & 3.97 & 185.3 \\
      \bottomrule
    \end{tabular}
\end{wraptable}

As described in Sec. \ref{sec:arch_design}, the prediction head can be configured to model the token distribution over different spatial extents. We further analyze the interplay between the size of prediction target ("window size") and the supervision density (number of tokens per sequence used for loss computation). We directly compares the two architectural variants of the prediction head. The global variant corresponds to a window size of $256$, where the bidirectional Transformer head predicts the entire image grid at each step. The chunk-based variant corresponds to smaller window sizes (e.g., $16$, $64$), where the cross-attention head predicts a localized region. To ensure a fair comparison of architectural trade-offs, we keep the total computational budget for supervision approximately constant, approximated by the product of window size $w$ and supervision density $n$.

\begin{figure}[t]
  \centering
  \includegraphics[width=1.0\linewidth]{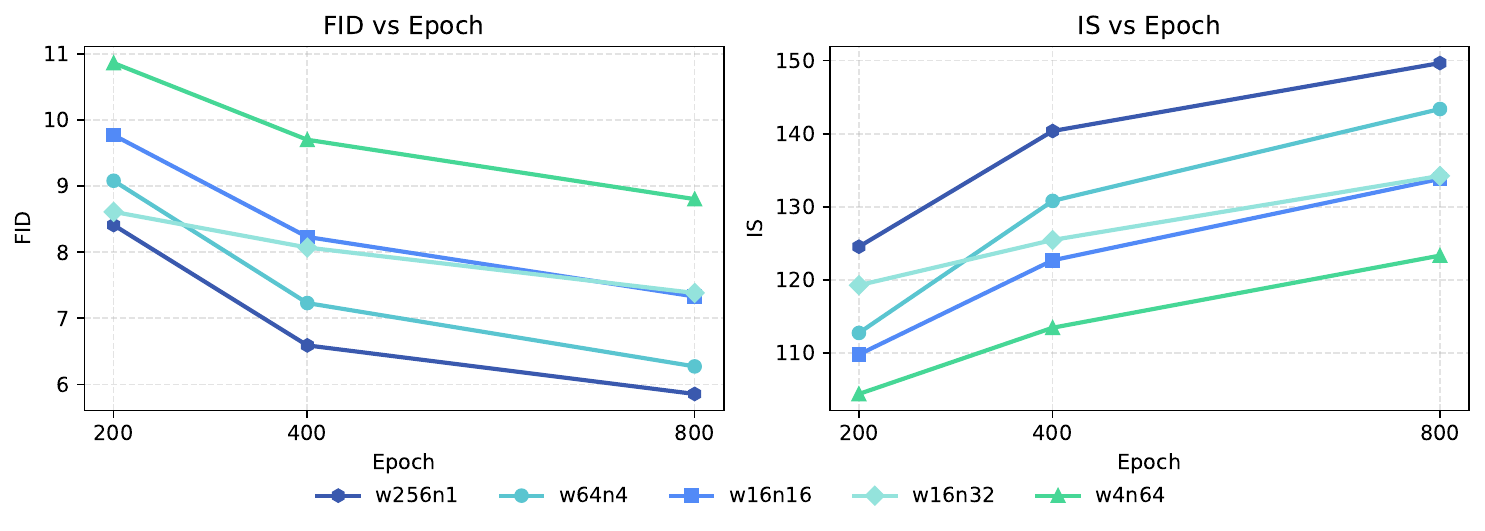}
  \caption{Effect of window size and supervision density on generation performance for Heptapod-B.}
  \label{fig:win_token_ablation}
\end{figure}

Figure \ref{fig:win_token_ablation} presents the FID and IS curves for five configurations. The results show a clear trend: models employing a larger prediction window consistently outperform those with a smaller window. The global variant ($w256n1$), which predicts the entire 256-token grid, achieves the best performance. In contrast, chunk-based models with smaller window size (e.g., $w16n16$, $w64n4$) yield worse performance, even when the supervision density ($n$) is increased to maintain a similar computational budget. While increasing $n$ within a moderately sized window ($w16n16$ vs.\ $w16n32$) can accelerate convergence, it does not surpass the full-window baseline and converges to similar final results.

These findings lead to a crucial conclusion: the spatial extent of the prediction target is more critical than the supervision density to learn long-term semantics. By compelling the model to predict distributions over the entire spatial context, the prediction head forces the model to capture long-range dependencies and holistic semantics. Restricting the prediction to smaller localized chunks, even with denser supervision, limits the global understanding and hinders the final generative quality. Based on this analysis, we adopt the global prediction head as the default for subsequent experiments. This ablation highlights the importance of maximizing the window size in the prediction objective and supports our core hypothesis that a holistic prediction task is key to effective visual language modeling.

\subsection{Benchmarking with Previous Generative Models}

\begin{table}[t]
  \caption{ImageNet $256\times256$ class-conditional generation \textbf{without} CFG. ``NAR'': non-autoregressive masked prediction. $\dagger$: uses an external SSL model. $^*$: results copied from \citet{chen2025toward}. ``Part'' for VAR indicates bidirectional attention within each scale (i.e., partially non-causal).}
  \label{table:main_results}
  \centering
  \small
  \begin{tabular}{c|c|c|c|cc}
    \toprule
    Type & Causal Attn. & Model & $\#$Params & FID$\downarrow$ & IS$\uparrow$  \\
    \midrule
     \multirow{6}{*}{Diffusion} & \multirow{6}{*}{No} & LDM-4~\cite{Rombach_2022_CVPR} & 400M & 10.56 & 103.5 \\ 
     & &  DiT-XL~\cite{Peebles_2023_ICCV} & 675M & 9.62 & 121.5  \\
     & & SiT-XL~\cite{10.1007/978-3-031-72980-5_2} & 675M & 8.30 & 131.7 \\ 
     & & REPA$^{\dagger}$~\cite{yu2024repa} & 675M & 5.90 & 157.8 \\
     & & MAETok$^{\dagger}$~\citep{chen2025masked} & 675M & 2.31 & 216.5 \\
     & & LightningDiT$^{\dagger}$~\cite{Yao_2025_CVPR} & 675M & 2.17 & 205.6 \\
     \midrule
     \multirow{3}{*}{NAR} & \multirow{3}{*}{No} & MaskGIT~\cite{Chang_2022_CVPR}  & 227M & 6.18 & 182.1  \\
     & & MAGVIT-v2~\cite{yu2024language} & 307M & 3.65 & 200.5  \\
     & & TiTok~\cite{NEURIPS2024_e91bf7df} & 287M & 4.44 & 168.2 \\ 
     \midrule
     \multirow{3}{*}{VAR} & \multirow{3}{*}{Part} & VAR-d20~\cite{NEURIPS2024_9a24e284} & 600M & 8.48$^*$ & 129.5  \\
     & & VAR-d24~\cite{NEURIPS2024_9a24e284} & 1.0B & 6.20$^*$ & 154.3  \\
     & & VAR-d30~\cite{NEURIPS2024_9a24e284} &2.0B & 5.26$^*$ & 175.6  \\
     \midrule
     \multirow{3}{*}{MAR} & \multirow{3}{*}{No}  & MAR-B~\cite{NEURIPS2024_66e22646} & 208M  & 3.48 & 192.4  \\
     & & MAR-L~\cite{NEURIPS2024_66e22646} & 479M & 2.60 & 221.4  \\
     & & MAR-H~\cite{NEURIPS2024_66e22646} & 943M & 2.35 & 227.8  \\ 
     \midrule
     \multirow{6}{*}{AR} & \multirow{6}{*}{Yes} &
     RQ-Transformer~\cite{Lee_2022_CVPR} & 1.4B & 8.71 & 119.0 \\
     & & RQ-Transformer~\cite{Lee_2022_CVPR}  & 3.8B & 7.55 & 134.0 \\
     & & LlamaGen-XL~\cite{sun2024autoregressive}  & 775M & 15.55 & 79.2 \\
     & & LlamaGen-XXL~\cite{sun2024autoregressive}  & 1.4B & 14.65 & 86.3  \\ 
     & & LlamaGen-3B~\cite{sun2024autoregressive} & 3B & 9.38 & 112.9 \\ 
     & & DiGIT$^{\dagger}$~\cite{NEURIPS2024_325ce329} & 732M & 3.39 & 205.96 \\ 
     \midrule
     \multirow{3}{*}{AR} & \multirow{3}{*}{Yes}
     & Heptapod-B (Ours)  & 208M  & 5.85 & 149.6 \\
     & & Heptapod-L (Ours)  & 478M  & 3.62 & 190.8  \\
     & & Heptapod-H (Ours) & 941M  & \textbf{2.70} & \textbf{229.8}  \\
    \bottomrule
  \end{tabular}
\end{table}

\begin{figure}[h]
  \centering
  \includegraphics[width=0.8\linewidth]{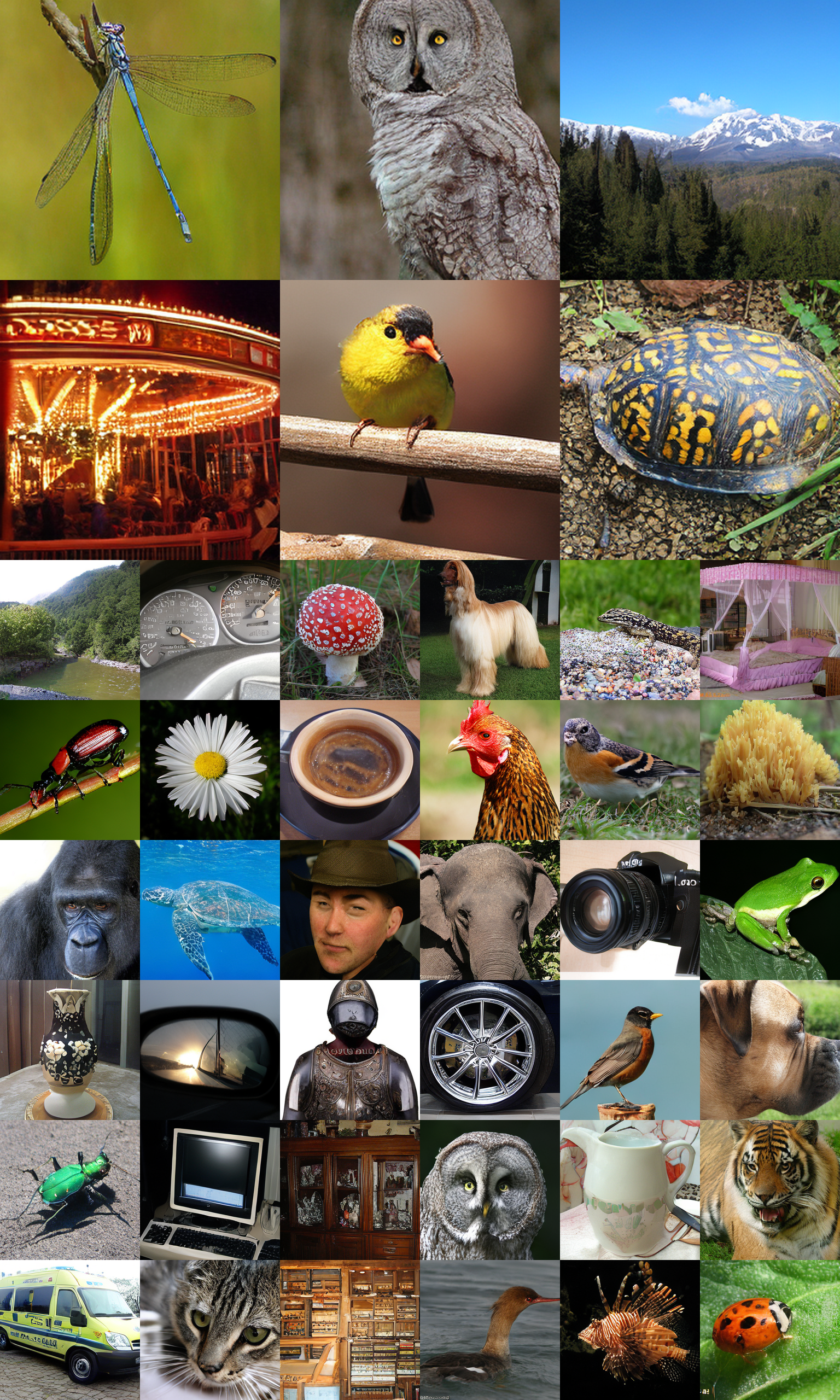}
  \caption{Example results generated by Heptapod-H.}
  \label{fig:case}
\end{figure}

We evaluate Heptapod against various generative models on ImageNet $256\times256$ conditional generation benchmark and the results are provided in Table \ref{table:main_results}. We train Heptapod with the $w256n1$ setting as described in Sec. \ref{sec:properties}. We provide generated examples in Fig. \ref{fig:case}.

A central goal of our work is to revitalize the causal autoregressive paradigm for vision. Compared to previous causal autoregressive models such as LlamaGen \cite{sun2024autoregressive}, Heptapod demonstrates a dramatic improvement. Heptapod-H achieves an FID of $2.70$ and an IS of $229.8$, decisively outperforming LlamaGen-3B (FID $9.38$, IS $112.9$) while using less than one-third of the parameters. These improvements underscore the efficacy of next 2D distribution prediction framework. By reformulating the learning objective, we enable a standard causal Transformer to learn rich visual semantics implicitly, a task where prior autoregressive models have struggled. 

Unlike many leading models (e.g., LightningDiT \cite{Yao_2025_CVPR}, MAETok \cite{chen2025masked}, DiGIT \cite{NEURIPS2024_325ce329}) that rely on external pre-trained SSL model to inject semantics into tokenizer, Heptapod uses a sole reconstruction-focused tokenizer. Moreover, all reported results are obtained without CFG, avoiding reliance on inference-time heuristics and highlighting intrinsic generative capability. This demonstrates that visual semantics can emerge from a well-posed generative objective, rather than being engineered into the tokenization. While MAR achieves better results through bidirectional attention, Heptapod with causal attention attains competitive results that approach MAR. The causal attention inherently operates at a lower computational cost compared to their bidirectional counterparts. In particular, Heptapod's adherence to causal attention ensures that it can be seamlessly integrated into a multimodal LLM, preserving the architectural coherence and deployment properties of language models.

These results support our central thesis that by returning to first principles and introducing a learning objective tailored to visual signals, a causal attention model trained without CFG and without semantic tokenizers can achieve strong performance. Heptapod thus offers a practical path for integrating visual generative training into the language modeling paradigm.

\subsection{Scale Supervision Density}

\begin{figure}[t]
  \centering
  \includegraphics[width=1.0\linewidth]{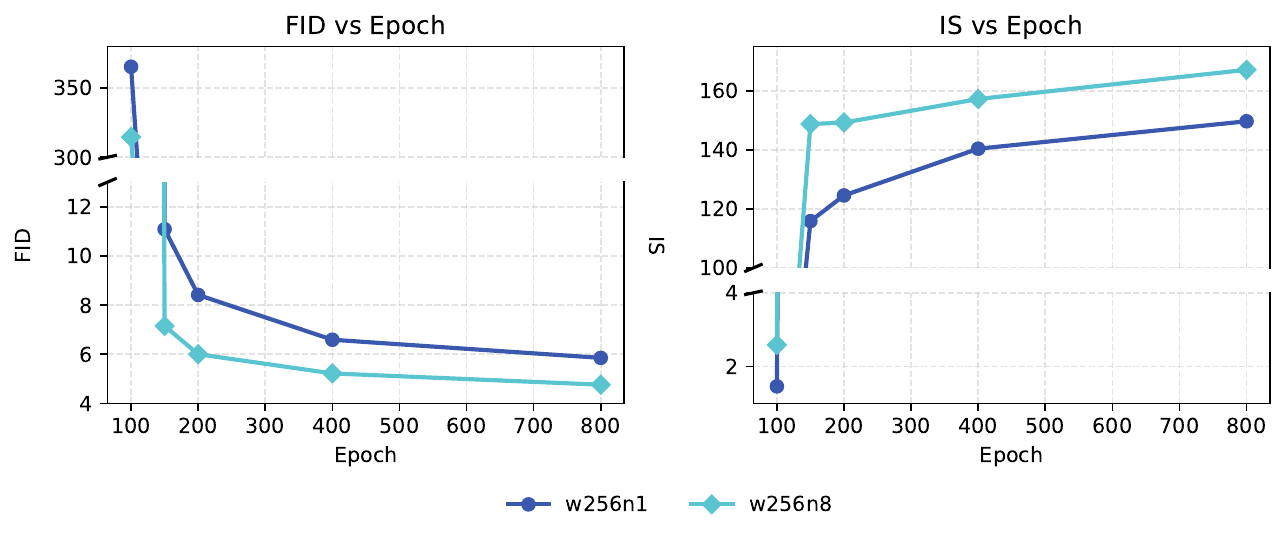}
  \caption{Effect of supervision density under a global prediction window ($w=256$). Increasing the number of tokens per sequence for loss computation (density $n$) consistently improves FID and IS. Results are obtained with Heptapod-B and CFG is disabled.}
  \label{fig:supervision_density}
\end{figure}

In Sec. \ref{sec:properties}, we observed that with a small prediction window (e.g., $w16$), simply increasing the number of tokens for loss computation from $16$ to $32$ ($w16n16$ vs.\ $w16n32$) yielded no improvement. This suggested that when the model's view is restricted to local context, denser supervision alone is insufficient to improve performance. We revisit this relationship under the global prediction window size of $256$, where the model is forced to predict the entire image. For example, $w256n1$ computes the loss on only a single token per sequence, whereas $w256n8$ increases this density eightfold. As shown in Fig. \ref{fig:supervision_density}, when the model is trained to predict the full 2D distribution, increasing supervision density leads to consistent improvements in both FID and IS metrics.

This contrast reveals a crucial insight: the effectiveness of dense supervision depends on the richness of the semantic signal. When the model is confined to small window size, denser supervision fails to provide a meaningful learning signal about global structure, hence negligible gains. 
With larger window size, the model must account for long-range dependencies and denser supervision provides richer gradients that improve both convergence and final quality. While scaling supervision density markedly accelerates convergence and improves final performance, it also increases the computational cost. Therefore, developing methods to enhance the training efficiency of our framework remains an important direction for future research.

\section{Discussion and Future Work}

\subsection{Connection with Multi-token Prediction}
There is a conceptual link between our framework and the multi-token prediction (MTP) method in LLMs \cite{pmlr-v235-gloeckle24a,liu2024deepseek}. MTP enriches the learning signal by training the model to predict several subsequent tokens at each step rather than just the immediate successor, which can mitigate exposure bias and accelerate training. In this context, our next 2D distribution prediction can be viewed as the extension of MTP to the visual domain: predicting the entire remainder of the token grid. This ``predict future N tokens'' objective is exceptionally suited for images, as it resolves the inherent ambiguity of ``next token'' in a non-sequential 2D space. The mandate to predict distributions for all future locations necessitates the learning of global structure and semantics, as local interpolation is no longer a sufficient strategy. We posit that this shift toward long-range prediction objective is a promising direction for language modeling in vision and beyond.

\subsection{Language Modeling on Acoustic Signals}
Our framework, especially the chunk-based prediction head described in Sec. \ref{sec:arch_design}, suggests a promising path toward language modeling on acoustic signals. Unlike images, audio can be extremely long or even unbounded in streaming settings, making the global prediction computationally infeasible. The chunk-based approach, which limits prediction to a finite local window, naturally aligns with this characteristic. Although our method is motivated by the emergence of visual semantics under holistic prediction, it is unclear whether the same principle would hold for acoustic signals. Addressing this question is a key direction for future work.

\subsection{Tokenizer in Heptapod}
Although our current implementation employs a visual tokenizer (such as VQ-VAE or VAE) to compress images, the core principle of our framework is fundamentally agnostic to the form of the visual tokens. The prediction head simply models distributions over the 2D spatial positions, regardless of whether those positions correspond to discrete tokens or continuous patches. Recent advances, such as Fractal Generative Models \cite{li2025fractal}, have demonstrated that it is possible to achieve high-fidelity image generation directly in pixel space through hierarchical modeling techniques, bypassing the need for tokenization in Fig. \ref{fig:intro1}.

\section{Conclusion}

In this work, we revisited the growing reliance on externally engineered semantics in visual language modeling and advocated for a return to the first principles that have driven the success of LLMs. By decoupling reconstruction from semantic learning and introducing the next 2D distribution prediction, we presented a novel framework that extends autoregressive modeling beyond the constraints of sequential data. Our approach encourages the model to develop a holistic understanding of images by predicting the 2D distribution, effectively bridging the strengths of both autoregressive generation and masked autoencoding paradigms. This perspective not only resolves the inherent ambiguities of next-token prediction in the visual domain, but also lays a foundation for unified models that integrate generative and understanding capabilities across modalities.

\clearpage

\bibliographystyle{plainnat}
\bibliography{main}

\clearpage

\clearpage

\beginappendix

\section{Model Configuration}
\label{appendix:configuration}

Table~\ref{tab:model_config} summarizes the three model scales for causal Transformer (Base, Large, Huge) used throughout our experiments. For continuous tokenizer, following MAR \cite{NEURIPS2024_66e22646}, we configure the diffusion-style denoising MLP head with \{6, 8, 12\} blocks and widths \{1024, 1280, 1536\} for the Base, Large, and Huge models, respectively.

Table~\ref{tab:model_ablation} investigates how to allocate depth between the causal Transformer and the 2D prediction head for Heptapod-L. A balanced split (16/16) achieves the best FID and IS. Shifting layers from the head to the backbone (24/8) degrades performance, and an extreme allocation (31/1) leads to a sharp drop in both FID and IS. These results indicate that the prediction head must retain sufficient depth to project the 1D causal context into 2D spatial distributions and to model position-wise interactions at scale.

\begin{table}[t]
\centering
\caption{Model Configuration for causal Transformer.}
\label{tab:model_config}
\begin{tabular}{l c c c}
\hline
 & \textbf{Base} & \textbf{Large} & \textbf{Huge} \\
\hline
Depth    & 12 & 16 & 20 \\
Hidden Size  & 768 & 1024 & 1280 \\
FFN Dim & 3072 & 4096 & 5120 \\
Attention Heads            & 12 & 16 & 16 \\
Head Dim                   & 64 & 64 & 80 \\
\hline
\end{tabular}
\end{table}

\begin{table}[t]
\centering
\caption{Ablation study on the configuration of prediction head for Heptapod-L with $w256n1$. We vary the depth of layers between the causal Transformer and the prediction head while keeping the total fixed at 32 (e.g., 16/16, 24/8, 31/1).}
\label{tab:model_ablation}
\begin{tabular}{l c c c}
\hline
Depth (Causal Transformer / Prediction Head) & FID & IS \\
\hline
31 / 1 & 14.74 & 98.15 \\
24 / 8 & 4.88 & 158.93  \\
16 / 16  & 3.62 & 190.8 \\
\hline
\end{tabular}
\end{table}

\section{Visualization of Tokenizer and Attention}
\label{appendix:similarity_grid}
\begin{figure}[t]
  \centering
  \includegraphics[width=1.0\linewidth]{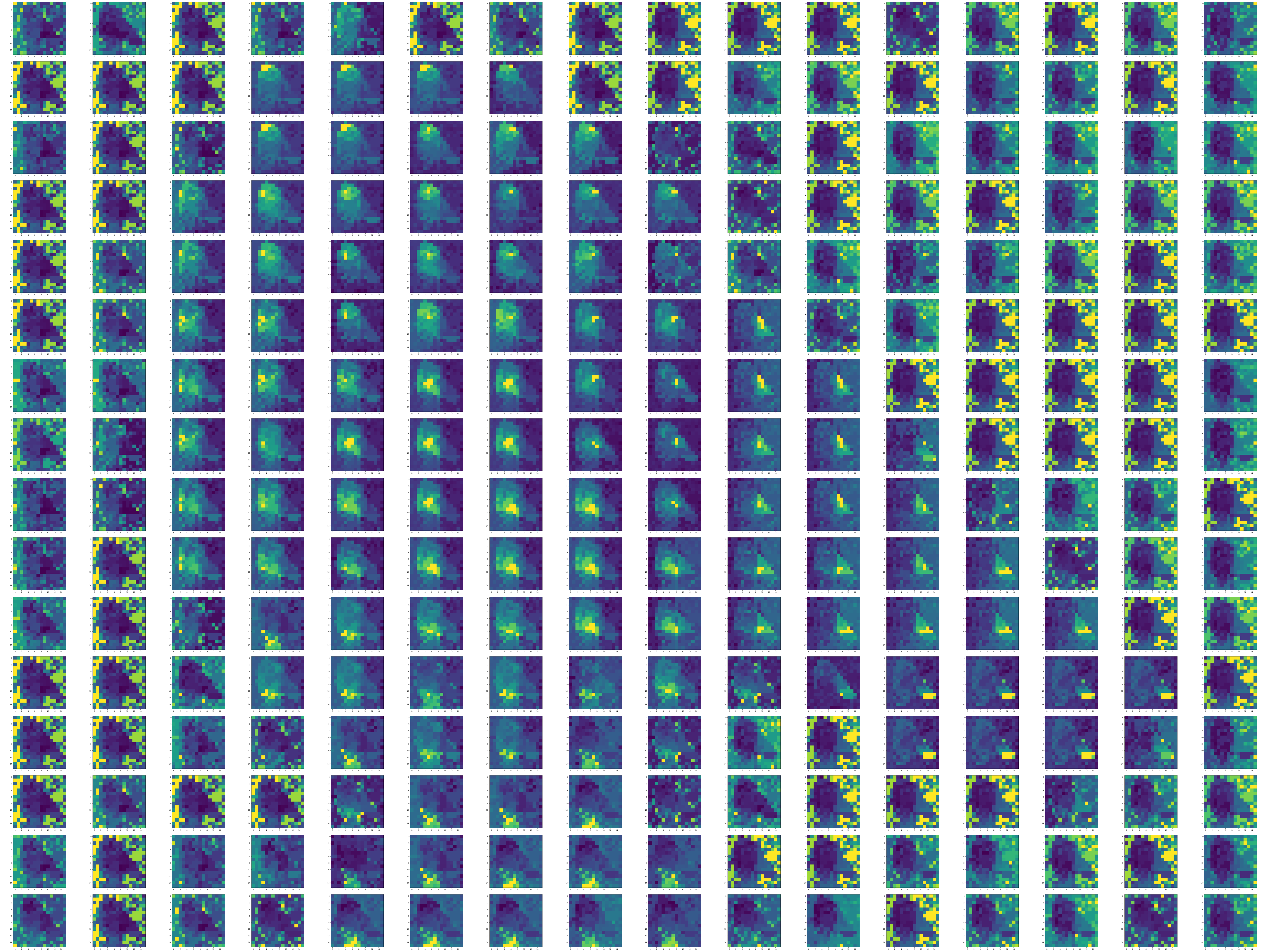}
  \caption{Visualization of semantic tokens with cosine similarity.}
  \label{fig:similarity_grid_digit}
\end{figure}

\begin{figure}[t]
  \centering
  \includegraphics[width=1.0\linewidth]{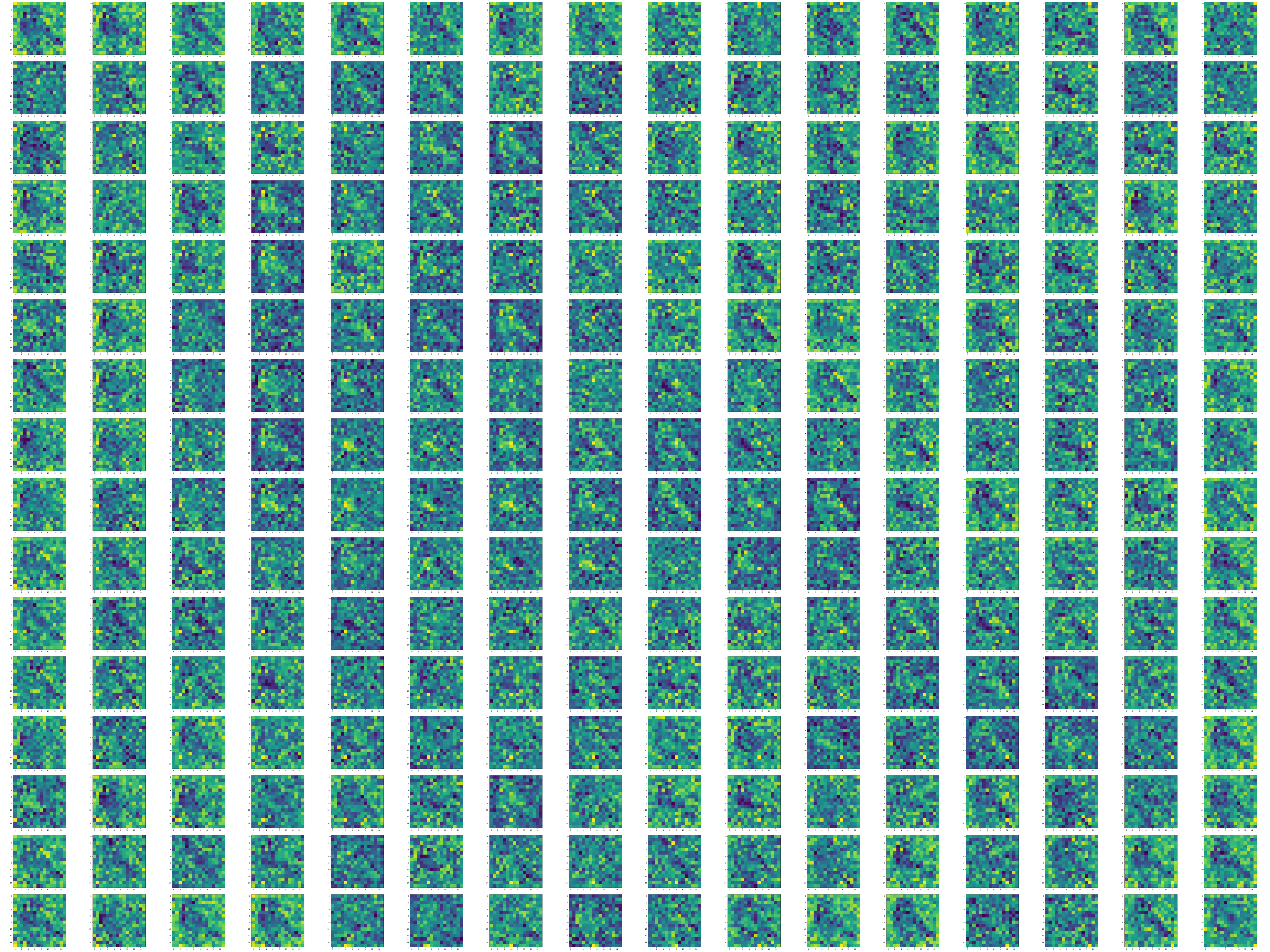}
  \caption{Visualization of VQ-VAE tokens with cosine similarity.}
  \label{fig:similarity_grid_vq}
\end{figure}

\begin{figure}[t]
  \centering
  \includegraphics[width=1.0\linewidth]{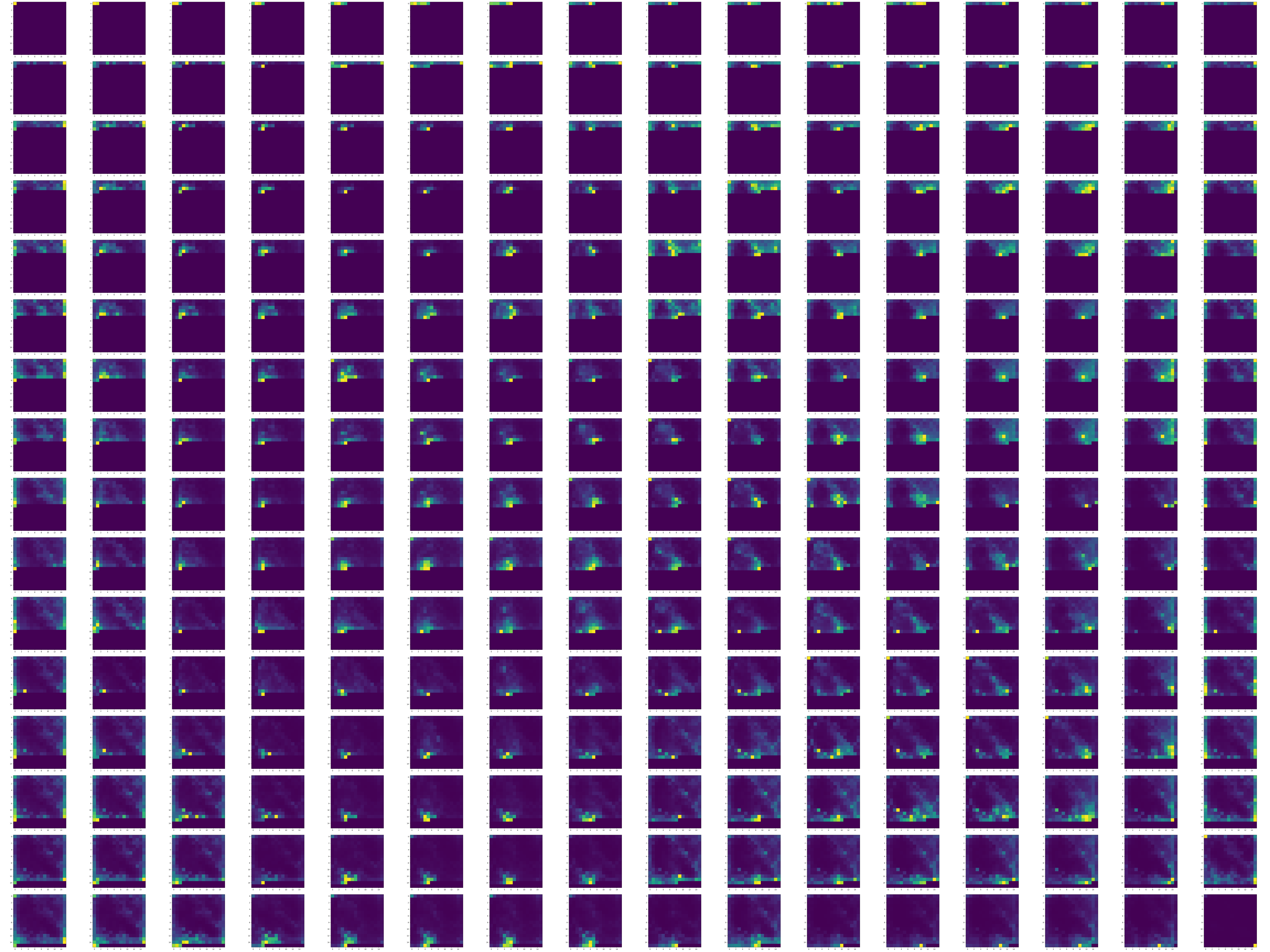}
  \caption{Visualization of attention maps with semantic tokenizer.}
  \label{fig:attention_grid_digit}
\end{figure}

\begin{figure}[t]
  \centering
  \includegraphics[width=1.0\linewidth]{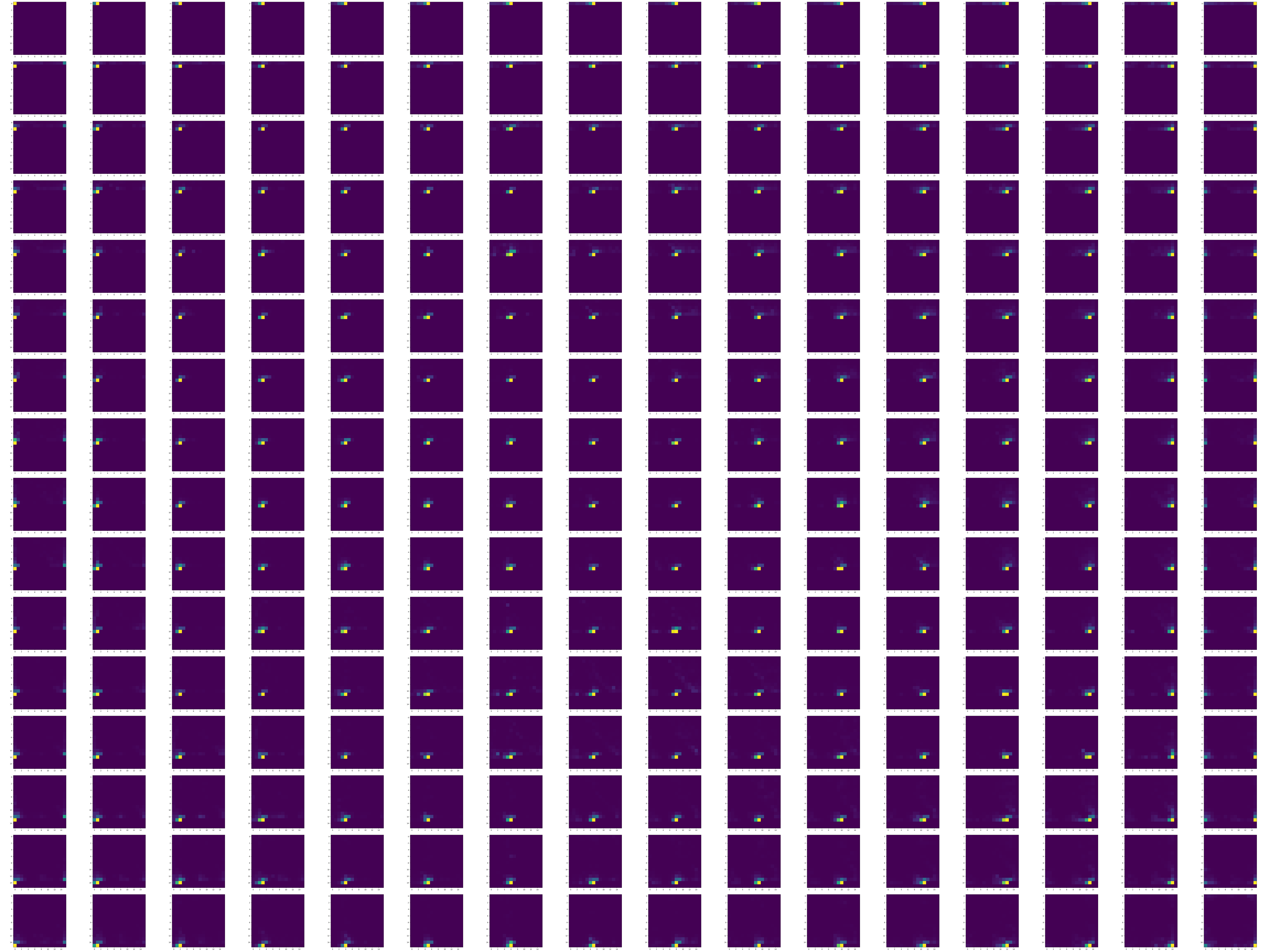}
  \caption{Visualization of attention maps with VQ-VAE tokenizer.}
  \label{fig:attention_grid_vq}
\end{figure}

\end{document}